\pdfoutput=1

\documentclass[11pt]{article}

\usepackage[final]{acl}

\usepackage{times}
\usepackage{latexsym}
\usepackage{booktabs}
\usepackage{graphicx}
\usepackage{adjustbox}
\usepackage{multirow}
\usepackage{subcaption}
\usepackage{amsmath}
\usepackage{amsfonts}
\usepackage{amssymb}
\usepackage{multirow}
\usepackage{bigdelim}
\usepackage[linesnumbered,ruled]{algorithm2e}

\usepackage[T1]{fontenc}

\usepackage[utf8]{inputenc}

\usepackage{microtype}

\usepackage{inconsolata}

\title{Looking Right is Sometimes Right: Investigating the Capabilities of Decoder-only LLMs for Sequence Labeling}

\author{David Dukić$^{\dagger}$ \and Jan Šnajder  \\ TakeLab, Faculty of Electrical Engineering and Computing, University of Zagreb \\ \texttt{\{david.dukic, jan.snajder\}@fer.hr}}

\begin{document}
\maketitle
\def\thefootnote{$\dagger$}\footnotetext{Corresponding author: \texttt{david.dukic@fer.hr}}
\renewcommand{\thefootnote}{\arabic{footnote}}
\begin{abstract}
Pre-trained language models based on masked language modeling (MLM) excel in natural language understanding (NLU) tasks. While fine-tuned MLM-based encoders consistently outperform causal language modeling decoders of comparable size, recent decoder-only large language models (LLMs) perform on par with smaller MLM-based encoders.  
Although their performance improves with scale, LLMs fall short of achieving state-of-the-art results in information extraction (IE) tasks, many of which are formulated as sequence labeling (SL). We hypothesize that LLMs' poor SL performance stems from causal masking, which prevents the model from attending to tokens on the right of the current token. 
Yet, how exactly and to what extent LLMs' performance on SL can be improved remains unclear. We explore techniques for improving the SL performance of open LLMs on IE tasks by applying layer-wise removal of the causal mask (CM) during LLM fine-tuning. This approach yields performance gains competitive with state-of-the-art SL models, matching or outperforming the results of CM removal from all blocks. Our findings hold for diverse SL tasks, demonstrating that open LLMs with layer-dependent CM removal outperform strong MLM-based encoders and even instruction-tuned LLMs.\footnote{\footnotesize{Code: \url{https://github.com/dd1497/llm-unmasking}.}} 


\end{abstract}

\section{Introduction}

Pre-trained language models (PLMs) built upon the Transformer architecture have demonstrated exceptional performance across many natural language understanding (NLU) tasks \citep{radford2018improving, devlin-etal-2019-bert, yang2019xlnet, clark2020electra, raffel2020exploring}. Typically, achieving state-of-the-art (SOTA) results in tasks such as sequence classification and sequence labeling involves a two-step process: pre-training on unlabeled corpora, followed by fine-tuning on task-specific data -- a process often referred to as transfer learning \citep{ruder-etal-2019-transfer, raffel2020exploring}. Two prevailing architectures emerged, each coupled with a compatible pre-training paradigm: (1) the decoder-only architecture, utilizing causal language modeling (CLM) for pre-training, and (2) the encoder-only architecture, with the masked language modeling (MLM) pre-training objective.

In transfer learning experiments that juxtapose models of a comparable number of parameters, MLM-based encoders consistently outperformed CLM-based decoders on NLU tasks \citep{devlin-etal-2019-bert}.\footnote{\footnotesize{Henceforth, we use the terms ``encoder'' and ``decoder'' to refer exclusively to MLM- and CLM-based encoder- and decoder-only variants, respectively.}} However, a shift in strategy emerged within the NLP community when encoder models ceased being scaled up to the same magnitude of parameters and pre-training data as their decoder counterparts. Consequently, there has been a pronounced trend toward scaling decoder models to multiple billion parameters, leading to a proliferation of large language models (LLMs). Combining LLM text generation capabilities with various prompting strategies can boost the performance on many NLU tasks, eliminating the need for fine-tuning model parameters \citep{liu2023pre}. 

\begin{figure}
    \centering
    \includegraphics[width=\columnwidth]{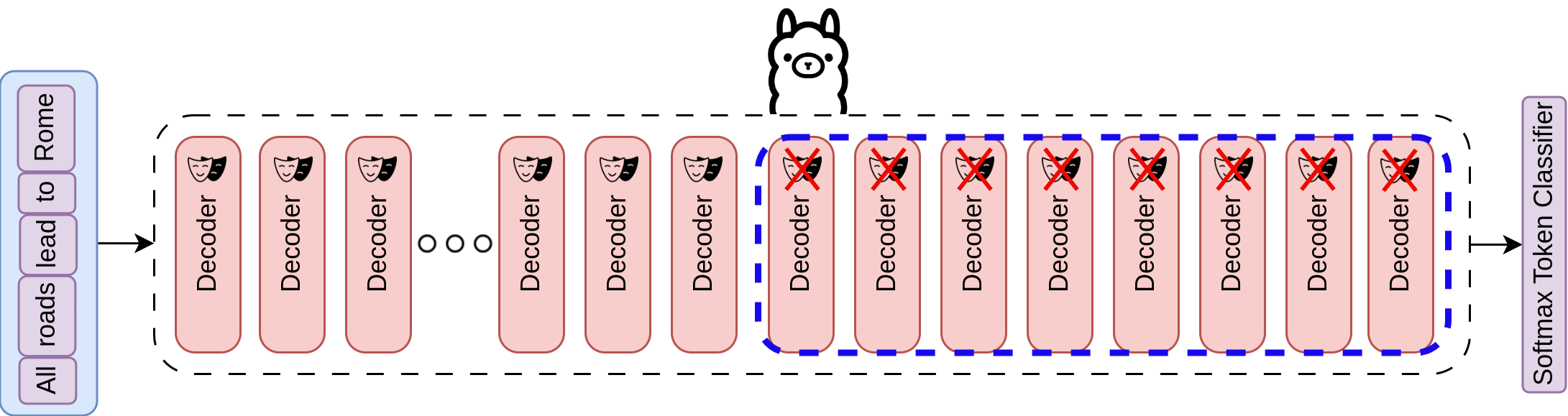}
    \caption{Layer-wise causal mask removal from decoder block groups in a decoder-only LLM. Here, the causal mask is removed from the top eight decoder blocks of the Llama2-7B model to enable bidirectionality during fine-tuning, which proves beneficial for many SL tasks.}
\end{figure}

Despite LLMs' good performance on NLU tasks, there is still considerable room for improvement, even for the largest decoder models, such as ChatGPT, which fall far behind SOTA results on fundamental NLP tasks. This holds particularly for information extraction (IE) tasks \citep{han2023information}, such as named entity recognition (NER), aspect-based sentiment analysis (ABSA), and event extraction (EE). These tasks are often formulated as sequence labeling (SL), and tackling SL by prompting LLMs proved quite difficult \citep{wang2023gpt}. However, it is unclear how exactly and to what extent the subpar performance of LLMs on SL tasks can be remedied.\footnote{\footnotesize{Although LLMs can solve many NLU tasks, including, but not limited to, IE tasks used for the knowledge base population, they have their limitations. We still need specialized methods for SL-based IE where we want to identify and count information elements in the text. This is especially useful for computational social science applications.}} As the community stopped scaling up encoders, LLMs are solidifying their position as the field's de facto standard. Thus, improving LLMs' SL performance is the sole viable option.

Although many SOTA LLMs are accessible only through paywalls, the community has responded by training and publicly releasing open LLMs with multiple billions of parameters, such as Llama2 \citep{touvron2023llama}, Mistral \citep{jiang2023mistral}, and OPT \citep{zhang2022opt}.  
Parameter-efficient fine-tuning (PEFT) techniques and quantization, including QLoRA \citep{dettmers2023qlora}, facilitate experimentation with fine-tuning these models on task-specific data. 
These strategies allow leveraging open LLMs' hidden states for task-specific classifiers, enabling fine-tuning on consumer-grade hardware with comparable or better performance than full fine-tuning. 

The poor performance of LLMs on SL tasks can be traced back to the information encoded in the hidden states. Using the hidden state of the decoder's last token serves as a reliable feature for classifying the input sequence. However, the same tactic does not suffice for SL because the causal mask (CM), integral to endowing decoders with text generation capabilities, limits bidirectional information flow, preventing the model from ``looking right,'' i.e., attending to tokens in the sequence positioned to the right of the current token. For many SL tasks, the token's label depends on the succeeding tokens, and omitting this context often results in subpar performance. Recently, \citet{li2023label} found that completely removing the CM from the Llama2 model during fine-tuning substantially enhances SL performance on NER. Others also experimented with the CM removal and enabling bidirectionality concurrently with our work \citep{behnamghader2024llm2vec,lee2024nv}.

In this paper, we explore techniques to enhance the SL performance of open LLMs on IE tasks. Building on insight from \citet{li2023label}, we hypothesize that removing the CM from all decoder layers may not be beneficial for all SL tasks. To confirm this, we experiment with layer-wise CM removal across LLM blocks, observing gains competitive with SOTA models, in contrast to removing or keeping the CM in all layers. We extend our analysis to a series of IE tasks (NER, ABSA, and EE) and demonstrate that layer-wise ``looking right'' is highly task-dependent. We compare against strong encoder-only sequence taggers and instruction-tuned LLMs. Open LLMs without CMs in specific layers outperform these baselines in almost all scenarios. 
Finally, we compare with encoder-only models, assessing the relative strengths of the decoder and encoder architectures regarding parameter scale, training data, and the particular SL task. 
We pre-train small encoder and decoder models from scratch with an identical number of parameters on the same data. 
Our findings indicate the superiority of the same-scale encoder architecture over the decoder on SL tasks, which is consistent even when the CM is removed while fine-tuning the small pre-trained decoders. This suggests that the observed SL performance gain in LLMs upon CM removal stems from scaling up. 

Our contributions are threefold: (1) We present evidence that layer-wise CM removal, followed by supervised fine-tuning, improves the performance of decoder-only LLMs on SL tasks compared to removing or keeping CM in all layers; (2) Layer-wise CM removal configuration that yields the highest gains strongly depends on the peculiarities of the SL task; (3) We show that the CM removal effect brings no gains on the small scale. We believe our results will contribute to building more performant decoder-only SL models.  


\section{Background and Related Work}

\paragraph{Encoders and Decoders.} 

The first encoder-only PLM, with MLM pre-training success, was BERT \citep{devlin-etal-2019-bert}, while the first decoder-only PLM with CLM pre-training that gained traction was GPT \citep{radford2018improving}. \citet{shoeybi2019megatron} showed that scaling the BERT model to four billion parameters with MLM and the GPT-2 model \citep{radford2019language} to eight billion parameters with CLM brings gains on NLU tasks. Nevertheless, the community stopped scaling up encoders and continued scaling decoders. This shift in strategy has given rise to notable properties in decoder-only LLMs, including enhanced zero- and few-shot learning capabilities \citep{wei2021finetuned, brown2020language}, the ability for in-context learning (ICL) \citep{brown2020language, liu2023pre}, and chain-of-thought prompting \citep{NEURIPS2022_9d560961}. Examples of LLMs with these emerging properties are GPT-3 \citep{brown2020language}, PaLM \citep{chowdhery2023palm}, and ChatGPT. 

Pre-training with CLM teaches the model to generate text and does not allow it to use the sequence context right to the token being processed. This is enabled via a CM, which prevents the model from attending to future tokens. While beneficial and necessary for coherent text generation, restricting the model to look right is detrimental if one wants to use the decoder to produce fully contextualized token-level embeddings with the decoder. 
Although decoder-only models perform poorly on sequence \emph{labeling} tasks (i.e., tasks where each individual token is assigned a label), they perform quite well on sequence \emph{classification} tasks (i.e., tasks where the entire sequence is assigned one label). 
Therefore, many decoder models implement sequence classification heads on top of decoder blocks but not SL heads \citep{wolf-etal-2020-transformers}. 

\citet{li2023label} found that removing the CM from all decoder layers of Llama2 \citep{touvron2023llama} improves the performance of named entity recognition (NER). However, \citet{li2023label} did not examine in depth why this phenomenon occurs, whether it occurs with layer-wise CM removal, and whether it persists for other IE tasks. We build on this and inspect the effect of different unmasking configurations across layers of open decoder-only LLMs. Furthermore, we train small decoders with CLM and small encoders with MLM to demonstrate that the superiority of unmasked LLMs is a consequence of the model and training data scale. 


\paragraph{Universal IE.}
Many IE tasks, such as NER, subtasks of ABSA, and EE, can effectively be formulated as SL tasks. Traditionally, these tasks were addressed with encoders, mapping each token's hidden state to task label space with a linear layer on top of the last encoder block \citep{pontiki-etal-2015-semeval, zhang-etal-2020-seqmix, zhou-chen-2021-learning}. However, this approach necessitates training a separate model for each task. More recent methods aim to infer the underlying structure of multiple IE tasks simultaneously and develop universal information extraction (UIE) models, employing encoder, decoder, or encoder-decoder architectures \citep{paolini2021structured, lu-etal-2022-unified, fei2022lasuie, wang-etal-2022-deepstruct, ping-etal-2023-uniex, zhu-etal-2023-mirror, lou2023universal, ding2024span}. 
While the practical advantages of UIE are obvious, including more IE tasks would still require re-training the UIE extractors. Therefore, we stick to the one model per SL task strategy, relying on parameter-efficient methods that yield compact and extensible models \citep{houlsby2019parameter}.

\paragraph{In-context Learning for IE.}
ICL is a powerful method for scrutinizing LLMs, employed by prompting LLMs with task demonstrations. IE tasks can be tackled with ICL, relying on the emergent properties of LLMs. Nevertheless, \citet{pang2023guideline} report that the performance of LLMs leveraging ICL lags behind the SOTA results of supervised IE models. They develop a model that learns to guide the prompt for improved LLM performance. Another notable example is GPT-NER \citep{wang2023gpt}, transforming the NER task into a generation one utilizing ICL and ChatGPT, while \citet{blevins-etal-2023-prompting} and \citet{mehta2024promptly} apply prompt-based methods to structured prediction tasks such as NER and semantic role labeling.

\paragraph{Instruction Tuning for IE.}
Releasing the weights of encoder-decoder and decoder-only pre-trained LLMs paved the way for instruction tuning (IT) \citep{mishra-etal-2022-cross}. This paradigm involves fine-tuning the model to address specific tasks in a supervised generative manner, providing the model with instructions for each data point. SL tasks can be adapted for compatibility with IT. Arguing that current prompt templates are primarily designed for sentence-level tasks and inappropriate for SL objectives, \citet{wang2022instructionner} reformulate NER as a generation problem relying on improved prompt templates and IT. 
Drawing on the fusion of UIE and IT, \citet{wang2023instructuie} reveal that multi-task IT can give results comparable to BERT in a supervised setting for NER. 
Finally, \citet{scaria2023instructabsa} find that instruction-tuned encoder-decoder PLMs excel at ABSA subtasks, leveraging the Tk-instruct model \citep{wang2022super}. 
Given that IT is a powerful supervised paradigm for combining LLMs' generation abilities with labeled data, we compare SL models against IT baselines.

\section{Layer Group Unmasking}

The CM is a crucial component of CLM-based decoders, preventing the model from attending to future tokens and facilitating autoregressive text generation. This constraint is enforced by defining the CM as a triangular matrix and adding this matrix to the dot product of the query and key attention matrices. The resulting sum is passed through the softmax function in the scaled dot-product attention mechanism, as introduced by \citet{vaswani2017attention}. Formally:
\begin{align*}
&\mathit{CM} = \begin{pmatrix}
    0 & -\infty & -\infty & \dots  & -\infty \\
    0 & 0 & -\infty & \dots  & -\infty \\
    \vdots & \vdots & \vdots & \ddots & \vdots \\
    0 & 0 & 0 & \dots  & 0
\end{pmatrix},\\\\
&\mathrm{Attn}(Q,K,V)=\mathrm{softmax}\left(\frac{QK^T + \mathit{CM}}{\sqrt{d_k}}\right)V.
\end{align*}

\noindent Here, $Q$, $K$, and $V$ are the query, key, and value attention matrices, respectively, and $d_k$ is the dimension of queries and keys. Effectively, applying softmax for tokens with value $-\infty$ in the CM results in attention scores being 0.

As reported by \citet{li2023label}, removing CM across all decoder layers of Llama2 during fine-tuning increases the model's performance on the NER SL task by a large margin. This is surprising because although, during pre-training, the CM mask was in place and the model was restricted from ``looking right,'' it learned to attend to future tokens by backpropagating on the training data over a few epochs with CM removed. Building on this and to allow the LLM to ``sometimes look right,'' we select a subset of decoder blocks for which we replace all $-\infty$ entries in the CM with zeros, effectively removing the CM (technically, we leave the positions of padding tokens to $-\infty$, but replace all others with 0). Since we group layers into groups of $b$ blocks, we refer to these decisions as layer group unmasking configurations. 

Considering that Llama2 is made up of $n=32$ decoder blocks, and each block can either remove or keep its CM, this gives a search space of $2^n=2^{32}$ possibilities. To reduce the search space to a manageable size, we group the $n$ decoder blocks into $m=4$ layer groups with $b=8$ consecutive decoder blocks per layer group and then jointly mask or unmask all blocks in each layer group. This leaves us with $2^m=16$ unmasking configurations per LLM and task. For ease of reference, we encode our unmasking configurations with binary four-digit codes ranging from 0000 (all four layer groups masked) to 1111 (all four layer groups unmasked), where each 0 and 1 denote a layer group that is masked or unmasked, respectively. Unmasking configurations are interpreted left to right (the first digit pertains to the layer group closest to the model's input).

\section{Experimental Setup} \label{sec:exp_setup}

\subsection{Sequence Labeling Tasks}

We focus our experiments on IE tasks framed as SL: NER, ABSA with aspect term extraction and polarity subtasks, and the trigger classification (TC) subtask of EE. We also include text chunking (shallow parsing), which, although not an IE task, is considered a prototypical SL task. Dataset statistics are shown in Table~\ref{tab:dataset_stats}. For each dataset, we calculate its \emph{right-side dependency relations ratio} (RDRR), defined as the ratio of right-side dependency relationships to the total number of left-side and right-side relationships counted for all labeled spans in the training set. Essentially, this metric indicates the degree to which labeled spans depend on the context to their right. We use spaCy \citep{spacy} dependency parser to obtain the dependency relations.

\begin{table}
  \centering
  \begin{adjustbox}{width=1.0\columnwidth}
  \small{\begin{tabular}{crrrrr}
    \toprule
    \multicolumn{1}{c}{\multirow{1}{*}{\textbf{Dataset}}} & \multicolumn{1}{c}{\textbf{Train}} & \multicolumn{1}{r}{\textbf{Validation}} & \multicolumn{1}{r}{\textbf{Test}} & \multicolumn{1}{r}{\textbf{Total}} & \multicolumn{1}{r}{\textbf{RDRR}}  \\
    \midrule
    \multirow{1}{*}{CoNLL03 NER} & \multirow{2}{*}{14041}  & \multirow{2}{*}{3250} & \multirow{2}{*}{3453} & \multirow{2}{*}{20744} & 0.593 \\
    \multirow{1}{*}{CoNLL03 Chunking} & & & & & 0.518 \\
    ACE05  & 14672 & 873 & 711 & 16256 & 0.411 \\     
    Rest14 & 2737 & 304 & 800 & 3841 & 0.397 \\
    \bottomrule
  \end{tabular}}
  \end{adjustbox}
  \caption{Statistics for the datasets and their splits. We show the number of sentences per split, the total number of sentences, and the right-side dependency relations ratio (RDRR).}
  \label{tab:dataset_stats}
\end{table}

\paragraph{NER and Text Chunking.}

For NER and text chunking, we choose CoNLL03, a standard and widely used benchmark \citep{tjong-kim-sang-de-meulder-2003-introduction}. We use the version from Hugging Face Datasets \citep{lhoest2021datasets} with an IOB2 sequence tagging scheme, which has a predefined train, validation, and test split.

\paragraph{Aspect Term Extraction and Polarity.}

We use the data from SemEval-2014 Task 4 and the restaurants domain \citep{pontiki-etal-2014-semeval} (Rest14). We merge the first two ABSA subtasks, aspect term extraction (ATE) and aspect term polarity (ATP), into one SL task (ATE+ATP). We tokenize the dataset with spaCy \citep{spacy} and match given character spans of aspect terms with token spans to obtain IOB2 tags (we discard 13 aspect terms that could not be matched in this way). Training and test split were predefined. Following prior work \citep{wang-etal-2021-automated}, we randomly sample 10\% of the sentences for the validation set.

\paragraph{Trigger Classification.} 

The ACE05 dataset \citep{doddington-etal-2004-automatic} is a widely used TC dataset. The TC task combines two EE tasks into a single SL task: trigger identification, i.e., finding spans of tokens constituting the event predicate and classifying them. We use the English train, validation, and test split obtained with the standard ACE pre-processing tool.\footnote{\footnotesize\url{https://bit.ly/ace2005-preprocessing}} We use this tool to obtain sentences and tokens and create IOB2 tags.

\paragraph{Evaluation.}

We evaluate all tasks with micro F1 score on IOB2 tag predictions with strict matching using seqeval \citep{seqeval}, where the predicted output span must exactly match the expected output span. We evaluate only the predictions on the first token of the tokenized words from the input sequence to obtain the same number of predictions as there are target IOB2 labels.

\paragraph{Instruction Tuning.}

To train LLMs using IT, each dataset needs to be further pre-processed to obtain instruction prompts. We form instructions similar to the ones used for NER by \citet{wang2022instructionner}, although we require a more strict output response from decoder-only LLMs, in line with the output format used by \citet{wang2023instructuie}. We create instruction prompts by parsing IOB2 tags to create desired outputs. See Appendix~\ref{sec:appendix_it} for instruction tuning examples. To ensure a fair evaluation consistent with models fine-tuned directly for SL, we heuristically map response spans of instruction-tuned models to IOB2 tags. We employ greedy span-based matching of predicted spans and their types with input tokens, similar to \citet{wang2022instructionner}. We treat all cases in which no predictions are made, or all predicted spans do not align with input tokens, or an exception arises during matching due to output generation stochasticity, as if the O tag was predicted for every input token. 

\begin{figure*}[!htb]
\begin{center}
\hspace*{\fill}%
\begin{subfigure}{0.25\textwidth}
    \centering
    \includegraphics[width=1.0\linewidth]{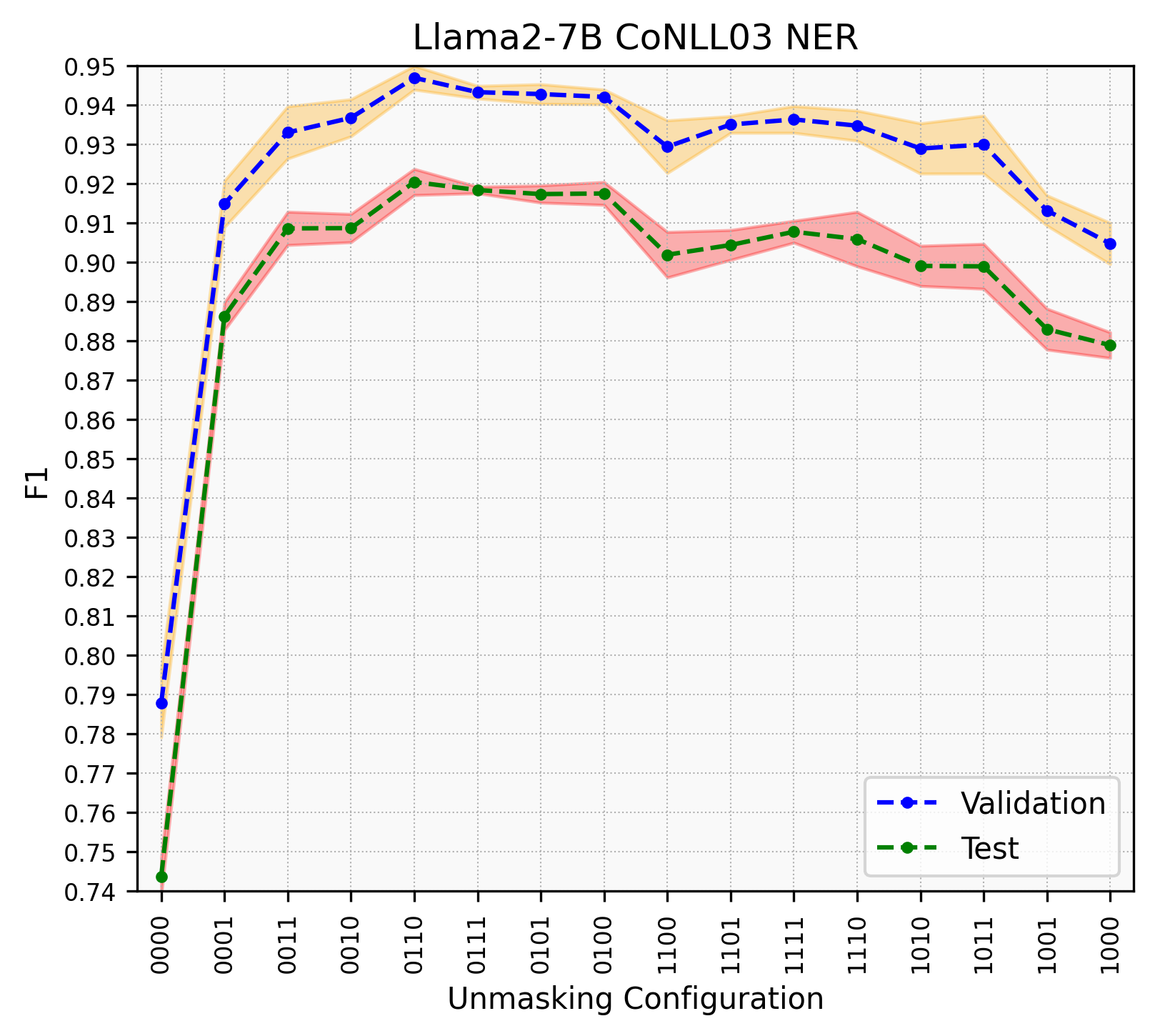} \\
    \includegraphics[width=1.0\linewidth]{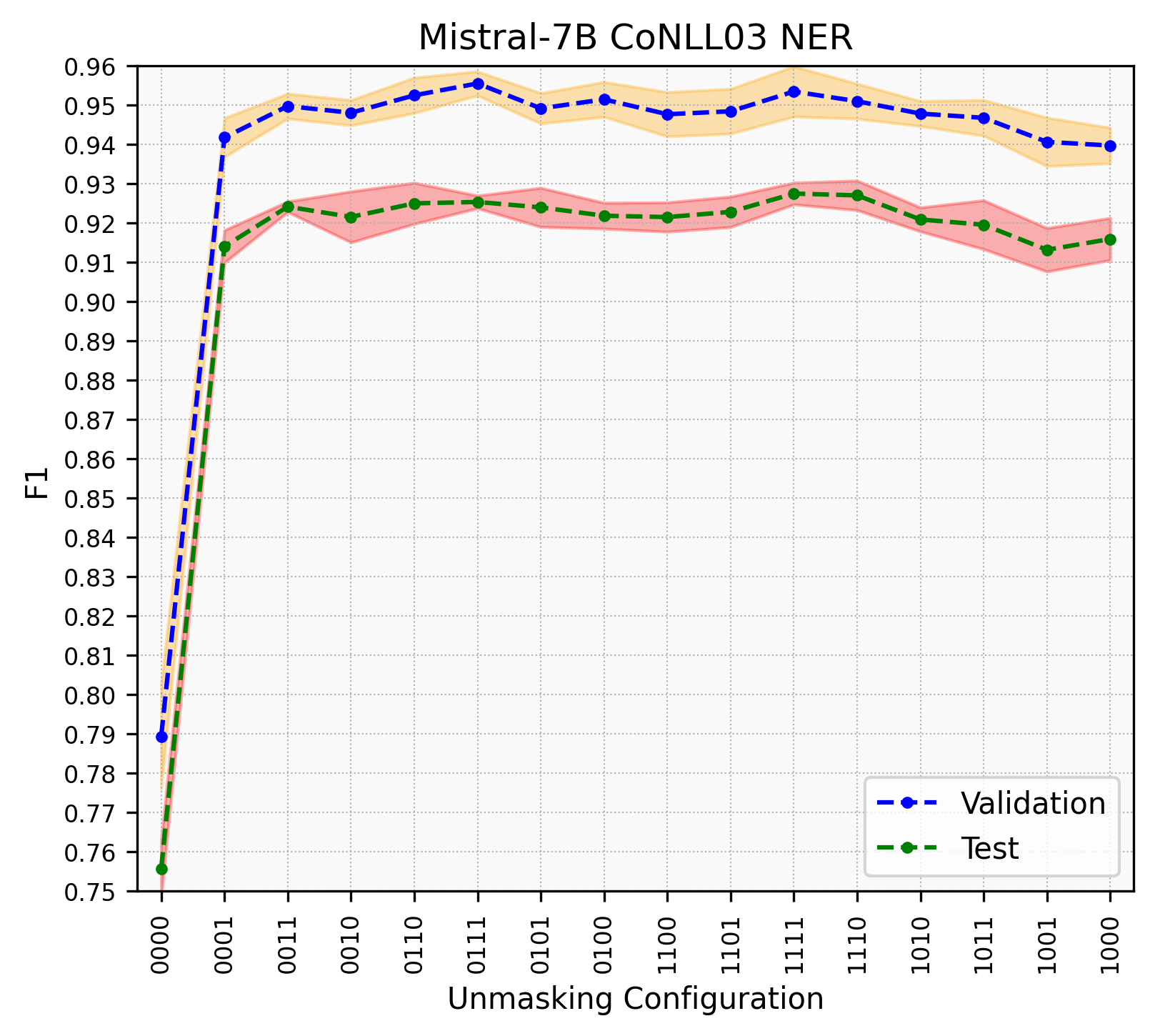}
    \caption{NER}
\end{subfigure}\hfill%
\begin{subfigure}{0.25\textwidth}
    \centering
    \includegraphics[width=1.0\linewidth]{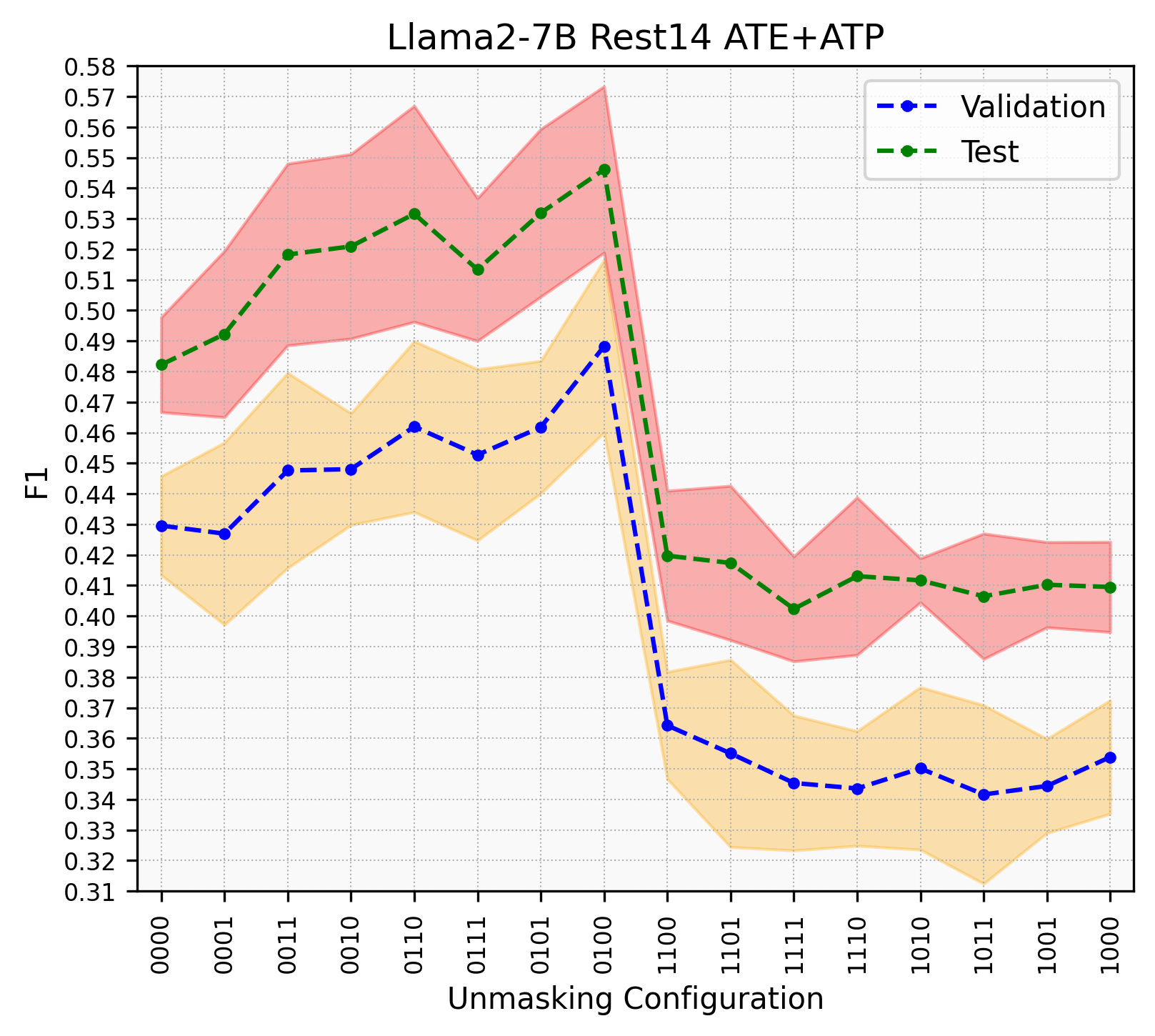} \\
    \includegraphics[width=1.0\linewidth]{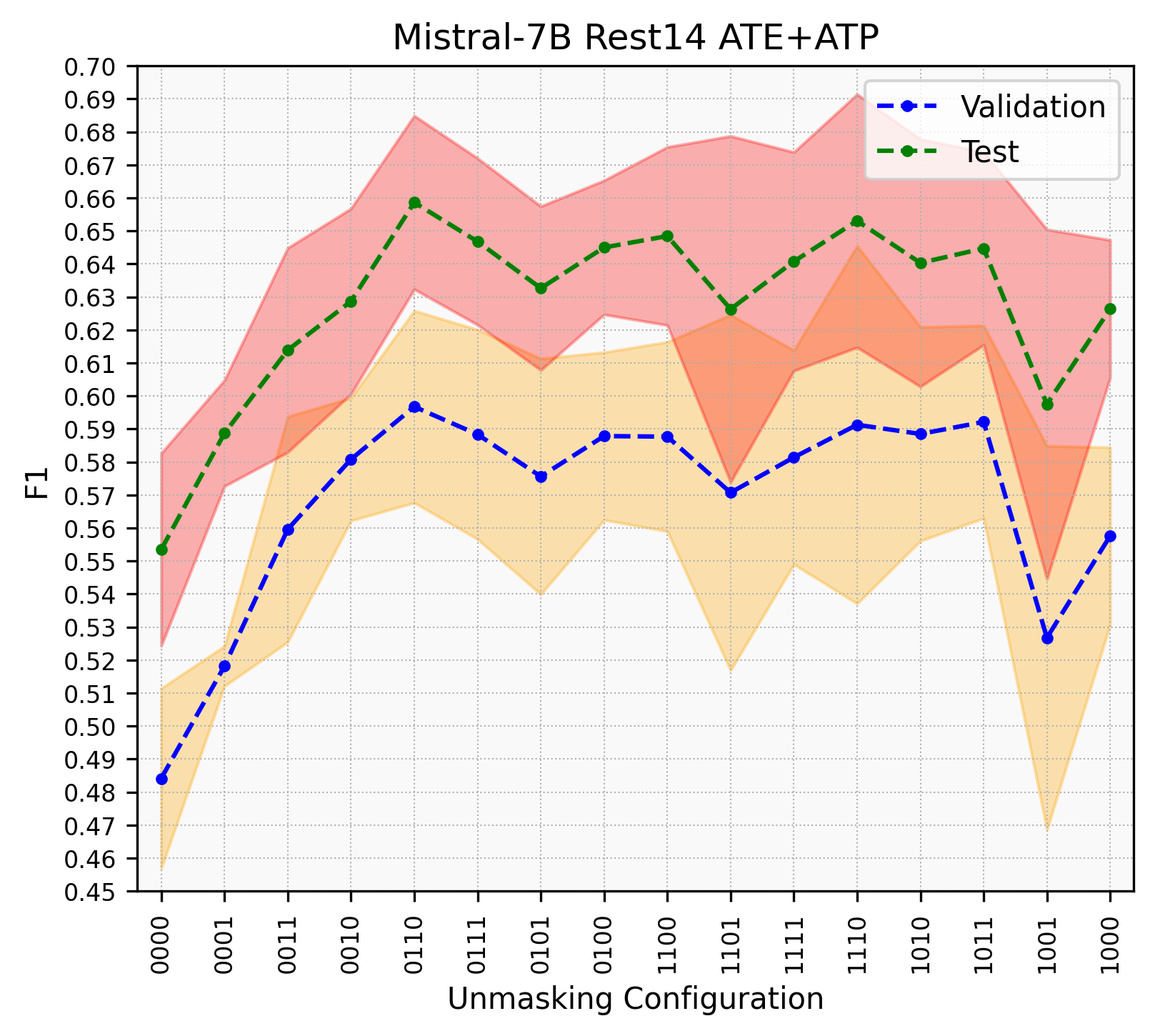}
    \caption{ATE+ATP}
\end{subfigure}\hfill%
\begin{subfigure}{0.25\textwidth}
    \centering
    \includegraphics[width=1.0\linewidth]{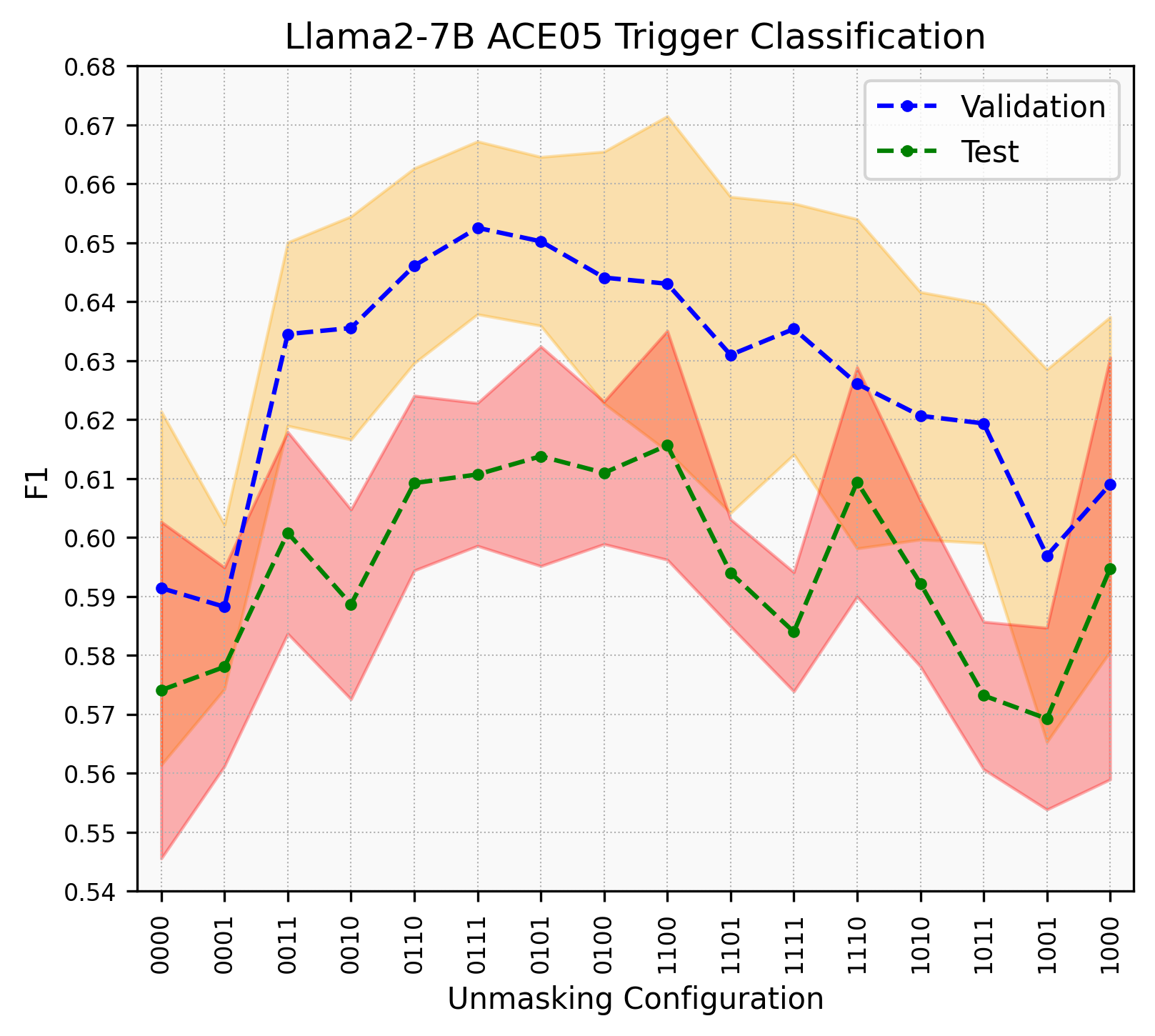} \\
    \includegraphics[width=1.0\linewidth]{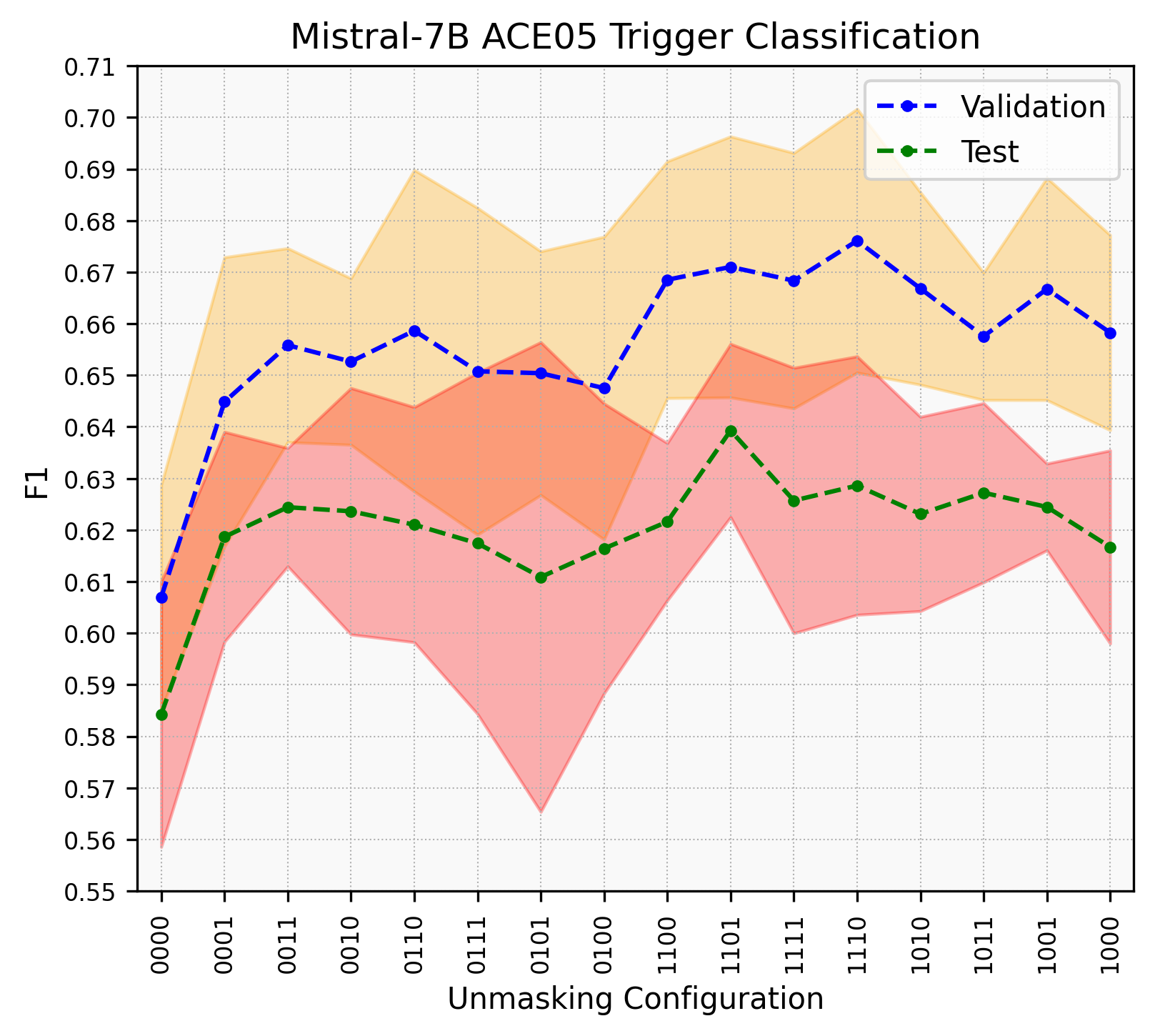}
    \caption{Trigger Classification}
\end{subfigure}\hfill%
\begin{subfigure}{0.25\textwidth}
    \centering
    \includegraphics[width=1.0\linewidth]{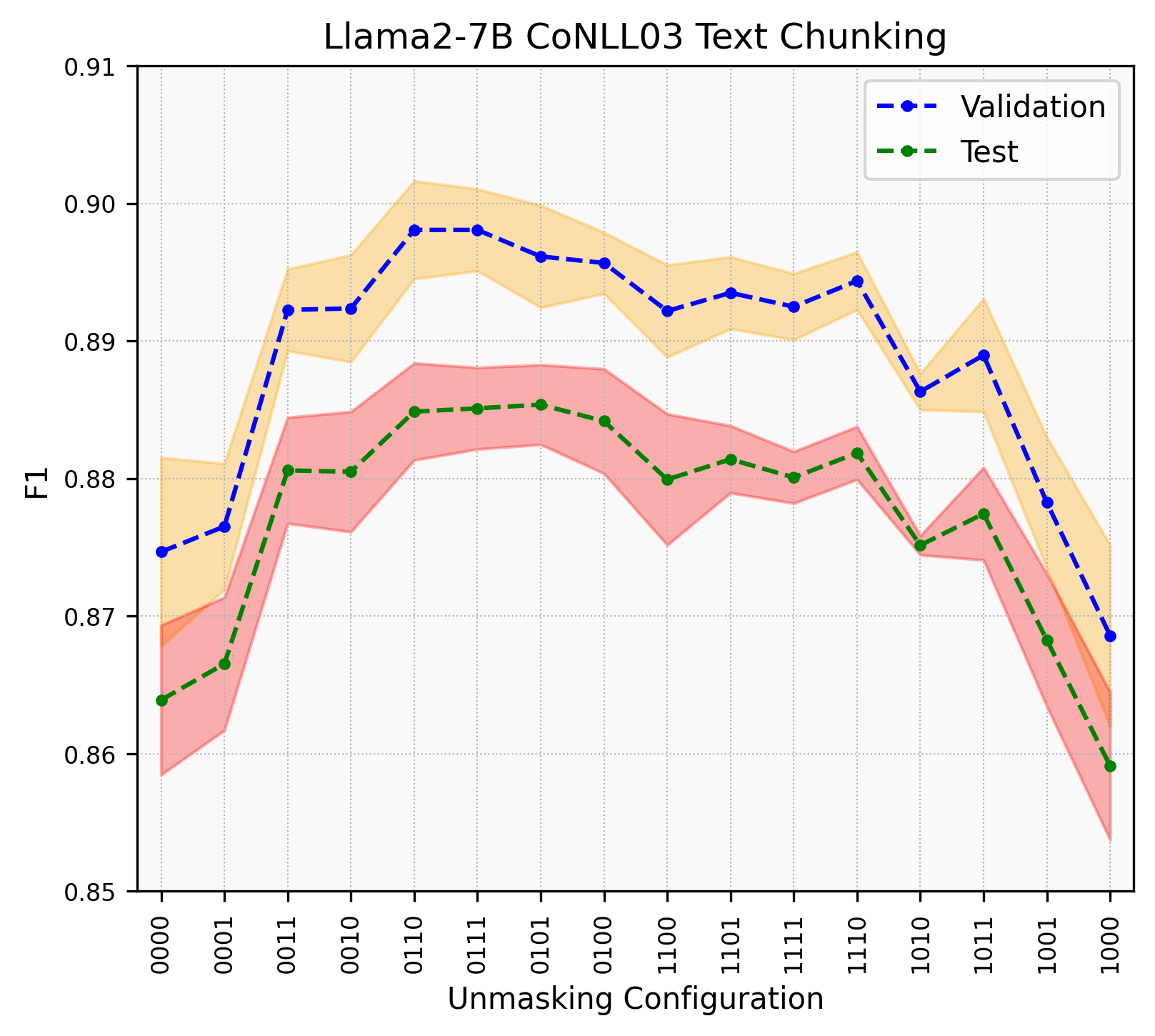} \\
    \includegraphics[width=1.0\linewidth]{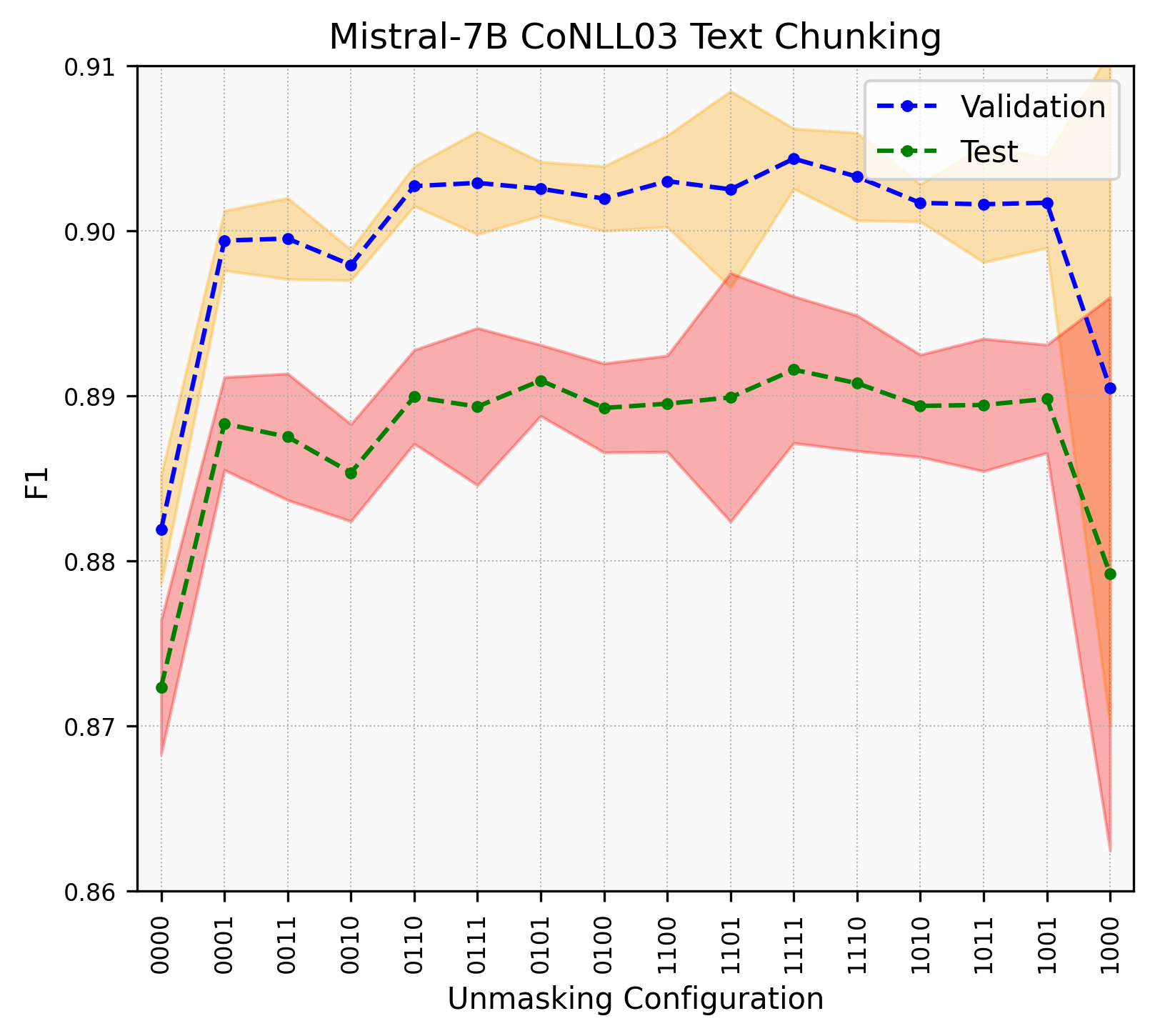}
    \caption{Text Chunking}
\end{subfigure}
\hspace*{\fill}%
\caption{Micro F1 SL scores with decoder-based LLMs and different unmasking configurations (sorted by Gray code starting with all decoder layers masked -- configuration 0000). Upper row plots show Llama2-7B model results, while lower row plots show Mistral-7B model results on validation and test sets of four SL tasks (left to right, one dataset per task). All results are averages of five runs. The shaded area corresponds to standard deviation.}
\label{fig:unmasking}
\end{center}
\end{figure*}

\subsection{Training Details and Hyperparameters}

\paragraph{Models.} We choose open LLMs for unmasking experiments and IT: Llama2-7B \citep{touvron2023llama} and Mistral-7B \citep{jiang2023mistral}, both with 7B parameters. For encoders, we use RoBERTa-base and RoBERTa-large PLMs with 125 and 355 million parameters, respectively \citep{liu2019roberta}. For SL experiments, we employ a classic softmax SL head on top of the pre-trained models. Model implementations and weights are taken from Hugging Face Transformers \cite{wolf-etal-2020-transformers}. If the SL head is not implemented, we add our implementation mimicking the one for RoBERTa models. See Appendix~\ref{sec:exp_setup_details} for more details.

\paragraph{Optimization and Pre-processing.} We use quantized LoRA (QLoRA) \citep{hu2021lora, dettmers2023qlora} for fine-tuning all models to ensure a fair comparison between the smallest and the largest models and enable fine-tuning under constrained computing resources. QLoRA is applied to query and value attention matrices inside each encoder or decoder block with a fixed rank of $r=64$, a scaling parameter of $\alpha=16$, and a dropout probability of $p=0.1$. 
This way, only decomposed query and value matrices, along with the SL head parameters, are optimized with cross-entropy loss, yielding a drastic trainable parameter reduction per model. The models are trained in bfloat16 precision, with loaded pre-trained weights in 4-bit NormalFloat data type, and we use double quantization. Using this setup, we were able to fit all models into 40GB of GPU memory of Ampere A100 and are trained with a consistent batch size of 16 per experiment. To use this batch size across models, we pre-process all datasets to a maximum tokenized sequence length of 128. This cutoff is optimal as there are <10 sentences for each dataset and split that end up truncated independent of the tokenizer used. 
We pad the sequences to the longest example in the batch and randomly sample examples for training depending on the seed. We train the models with paged 8-bit AdamW \citep{loshchilov2017decoupled} optimizer to handle the memory spikes \cite{dettmers2023qlora}. The parameters of AdamW are fixed to $\beta_1=0.9, \beta_2=0.95, \epsilon=1\mathrm{e}{-5}, \lambda=0.1$. For the learning rate scheduler, we choose cosine annealing scheduler \citep{loshchilov2016sgdr}. We apply gradient clipping set to $1.0$, gradient accumulation with four steps, and gradient checkpointing. We use a consistent learning rate of $2\mathrm{e}{-4}$ across all experiments and fine-tune models over a fixed number of five epochs. All results are averages of five runs with different seeds, and we always pick the last model for each seed.

\paragraph{Instruction Tuning Details.} We use Llama2-7B and Mistral-7B with CLM heads for IT. Training is done with QLoRA based on cross-entropy loss. All hyperparameters are inherited from SL experiments except for the batch size. The batch size depends on the maximum sequence length for packing the dataset examples. Short examples are packed together in the same input to increase training efficiency. This length is set to 512 for the Rest14 dataset and 1024 for other datasets since Rest14 has shorter instruction examples on average and has the smallest training set (cf.~Table~\ref{tab:dataset_stats}). We observe that ATE+ATP models need double the epochs (10 epochs) to learn to extract aspect terms and their polarity, and we hypothesize that this is due to the low number of training examples compared to other datasets (\textasciitilde3k vs.~\textasciitilde14k).


\section{Results}

\subsection{Layer Group Unmasking Results}

Figure~\ref{fig:unmasking} shows the validation and test evaluation sets performance of Llama2-7B and Mistral-7B for different unmasking configurations and four SL tasks. We observe a number of interesting phenomena: (1) In most cases, removing the CM in as little as one layer group significantly boosts the F1 score compared to keeping the CM in all layers (configuration 0000); (2) It is rarely the case that removing CM from all layer groups (configuration 1111), compared to removing it only from some layer groups, yields the highest score on evaluation sets; (3) Depending on the task, the boosts from the CM removal can vary (highest for NER, lowest for chunking) and deviate from minimal (NER) to substantial (TC); (4) Mistral-7B shows superior performance over Llama2-7B model across tasks and best configurations; (5) The shapes of validation and test curves show high overlap across configurations for a fixed model and dataset.

Unmasking gains are the highest for NER, and the results are the most consistent for CoNLL03 data (NER and chunking tasks). This aligns perfectly with the values of the RDRR (cf.~Section~\ref{sec:exp_setup}), suggesting that labeled NER spans depend more on the right context than on the left (cf.~Table~\ref{tab:dataset_stats}). We generally observe that unmasking layers closer to the model's output yields higher gains than unmasking layers closer to the model's input, except for ATE+ATP and TC tasks. Such a regularity would align with the finding from the literature that higher layers in neural language models are more localized and task-specific \citep{wu-etal-2020-similarity}. To verify whether such a regularity generally holds, we conduct a one-sided two-sample t-test for the difference of F1 score means on validation and test sets per task. We compare the means of each unmasking configuration, which contains at least one unmasked layer group, with all other possible combinations where at least one additional layer group is unmasked closer to the model's output. For example, we compare the F1 score of unmasking configuration 0100 to configurations \{0100, 0101, 0110, 0111\}. We repeat the same procedure for the following configurations: 0100, 0110, 1000, 1010, 1100, and 1110. The differences are significant for NER and chunking evaluation sets ($p<.01$) but not significant for ATE+ATP and TC evaluation sets ($p>.01$). Applying CM removal to individual layers as opposed to layer groups could give even larger F1 score boosts. The performance of Llama2-7B on the ATE+ATP task displays an unusual pattern where better performance is achieved when CM is preserved in all layer groups, as opposed to its complete removal.  
Here, allowing the model to ``look right'' in all layers hurts the performance, which aligns with its low RDRR value. This trend is not present for any other task and model combination. Furthermore, the Rest14 dataset is the only one where higher scores were achieved on the test set than on the validation set. This could be attributed to taking a random sample of training data as a validation set. Micro F1 score deviations across evaluation sets are the largest on ATE+ATP and TC tasks, possibly due to the overall low number of examples per evaluation set for Rest14 and ACE05 datasets (<1000).

\begin{table*}[!htb]
\centering
\begin{adjustbox}{width=1.0\linewidth}
\small{\begin{tabular}{l|cccccccccc}
\toprule
\multicolumn{3}{c}{\multirow{1}{*}{\textbf{Model}}} & \multicolumn{2}{c}{\textbf{CoNLL03 NER}} & \multicolumn{2}{c}{\textbf{Rest14 ATE+ATP}} & \multicolumn{2}{c}{\textbf{ACE05 Trigger Clf.}} 
& \multicolumn{2}{c}{\textbf{CoNLL03 Chunking}}  \\
\midrule
\multicolumn{3}{c}{\multirow{2}{*}{Llama2-7B-SL}}     &  Valid F1 & Test F1 &  Valid F1 & Test F1 &  Valid F1 & Test F1 &  Valid F1 & Test F1  \\
\cmidrule(lr){4-4} \cmidrule(lr){5-5} \cmidrule(lr){6-6} \cmidrule(lr){7-7} \cmidrule(lr){8-8} \cmidrule(lr){9-9} \cmidrule(lr){10-10} \cmidrule(lr){11-11} \multicolumn{3}{c}{} & \multicolumn{2}{c}{$\rho=0.999$} & \multicolumn{2}{c}{$\rho=0.994$} & \multicolumn{2}{c}{$\rho=0.806$} & \multicolumn{2}{c}{$\rho=0.998$} \\
\cmidrule(lr){1-3} \cmidrule(lr){4-5} \cmidrule(lr){6-7} \cmidrule(lr){8-9} \cmidrule(lr){10-11}
\multirow{7}{*}{\rotatebox[origin=c]{90}{\shortstack{Unmask Config.}}} 
& & \multicolumn{1}{l}{0000} & 0.788 & 0.744 & 0.430 & 0.482 & 0.591 & 0.574 & 0.875 & 0.864 \\
& & \multicolumn{1}{l}{1111} & 0.936 & 0.908 & 0.345 & 0.402 & 0.635 & 0.584 & 0.892 & 0.880 \\
\cmidrule(lr){2-3}
& \multirow{5}{*}{\rotatebox[origin=c]{90}{\shortstack{Best}}} & \multicolumn{1}{l}{0100} & -- & -- & 0.488 & 0.546 & -- & -- & -- & -- \\
& & \multicolumn{1}{l}{0101} & -- & -- & -- & -- & -- & -- & -- & 0.885 \\
& & \multicolumn{1}{l}{0110} & 0.947 & 0.920 & -- & -- & -- & -- & 0.898 & -- \\
& & \multicolumn{1}{l}{0111} & -- & -- & -- & -- & 0.653 & -- & -- & -- \\
& & \multicolumn{1}{l}{1100} & -- & -- & -- & -- & -- & 0.616 & -- & -- \\
\cmidrule(lr){2-3}
\multicolumn{3}{c}{\multirow{2}{*}{Mistral-7B-SL}}         &  Valid F1 & Test F1 &  Valid F1 & Test F1 &  Valid F1 & Test F1 &  Valid F1 & Test F1  \\
\cmidrule(lr){4-4} \cmidrule(lr){5-5} \cmidrule(lr){6-6} \cmidrule(lr){7-7} \cmidrule(lr){8-8} \cmidrule(lr){9-9} \cmidrule(lr){10-10} \cmidrule(lr){11-11} \multicolumn{3}{c}{} & \multicolumn{2}{c}{$\rho=0.999$} & \multicolumn{2}{c}{$\rho=0.983$} & \multicolumn{2}{c}{$\rho=0.899$} & \multicolumn{2}{c}{$\rho=0.994$} \\
\cmidrule(lr){1-3} \cmidrule(lr){4-5} \cmidrule(lr){6-7} \cmidrule(lr){8-9} \cmidrule(lr){10-11}
\multirow{6}{*}{\rotatebox[origin=c]{90}{\shortstack{Unmask Config.}}} 
& & \multicolumn{1}{l}{0000} & 0.789 & 0.756 & 0.484 & 0.553 & 0.607 & 0.584 & 0.882 & 0.872 \\
& & \multicolumn{1}{l}{1111} & 0.953 & \textbf{0.927} & 0.581 & 0.641 & 0.668 & 0.626 & \textbf{0.904} & \textbf{0.892} \\
\cmidrule(lr){2-3}
& \multirow{4}{*}{\rotatebox[origin=c]{90}{\shortstack{Best}}} & \multicolumn{1}{l}{0110} & -- & -- & 0.597 & 0.659 & -- & -- & -- & -- \\
& & \multicolumn{1}{l}{0111} & \textbf{0.956} & -- & -- & -- & -- & -- & -- & -- \\
& & \multicolumn{1}{l}{1101} & -- & -- & -- & -- & -- & \textbf{0.639} & -- & -- \\
& & \multicolumn{1}{l}{1110} & -- & \textbf{0.927} & -- & -- & 0.676 & -- & -- & -- \\
\cmidrule(lr){2-3}
\midrule
\multicolumn{3}{c}{RoBERTa-base-SL}  & 0.897 & 0.883 & 0.313 & 0.369 & 0.609 & 0.508 & 0.889 & 0.877     \\
\multicolumn{3}{c}{RoBERTa-large-SL} & 0.924 & 0.900 & 0.403 & 0.474 & \textbf{0.698} & 0.628 & 0.891 & 0.877    \\
\multicolumn{3}{c}{Llama2-7B-IT}     & 0.778 & 0.771 & 0.523 & 0.608 & 0.375 & 0.347 & 0.833 & 0.818   \\
\multicolumn{3}{c}{Mistral-7B-IT}    & 0.897 & 0.887 & \textbf{0.646} & \textbf{0.733} & 0.477 & 0.461 & 0.873 & 0.860  \\
\bottomrule
\end{tabular}}
\end{adjustbox}
\caption{Validation and test micro F1 SL scores for Llama2-7B and Mistral-7B models with various unmasking configurations (0000, 1111, and other configurations which surpass the F1 score of 1111 over SL datasets -- denoted as Best) are in the upper table part. The results for SL encoders and IT baselines are in the lower table part. The best results by dataset and evaluation set are in \textbf{bold}. For Llama2-7B and Mistral-7B, we report the Pearson correlation coefficient between validation and test unmasking configurations ($\rho$). All results are averages over five runs.}
\label{tab:comparison}
\end{table*}

\subsection{Comparison with Baselines}

Table~\ref{tab:comparison} shows the performance scores of the best unmasking configurations and the strong encoder and IT baselines. We compare against RoBERTa PLMs fine-tuned for SL task and instruction-tuned Llama2-7B and Mistral-7B LLMs. The standard 0000 and 1111 configurations are compared with the best configurations per model and task. We also report the Pearson correlation coefficient $\rho$ between all validation and test unmasking configurations. For all SL tasks and the Llama2-7B model, we observe a consistent improvement of best configurations over configuration 1111. Similar holds for Mistral-7B, except for chunking, where 1111 yields the highest F1 scores. Conforming to the findings from \citet{scaria2023instructabsa}, IT is most beneficial for the ATE+ATP task, outperforming any unmasking configuration on evaluation sets. RoBERTa-large surpasses all other models on the ACE05 validation set but fails to do so on the test set. 
RoBERTa baselines achieve high scores on SL tasks, except for ATE+ATP. Training with QLoRA combined with layer group unmasking creates high-performing and compact models, requiring a small number of additional parameters trained for each SL task. 
Correlations between configurations on evaluation sets are high for each task, while models trained on ACE05 exhibit the lowest $\rho$. A high overall $\rho$ indicates that the optimal unmasking configuration can be determined using the validation set.

\subsection{Comparison with SOTA IE Models}

Our results are competitive with SOTA. However, a fair comparison is challenging due to differences in training and evaluation. For example, the SOTA F1 score for CoNLL03 NER, reported by \citet{wang-etal-2021-automated}, is $0.946$. This result, however, was obtained using a model trained on the merged training and validation sets. Strong results on CoNLL03 NER were reported by \citet{liu-etal-2022-autoregressive}, reaching an F1 score of $0.941$ without task-specific feature engineering, relying solely on a conditional language model with explicit modeling of the target structure. Further, the authors whose work we build upon, \citet{li2023label}, report SOTA results on CoNLL03 NER. The score they report is $0.932$, which is competitive, but not SOTA. Upon code inspection, we found that they truncate sequences longer than 64 tokens to fit the data into GPU memory. In contrast, we managed to keep the sequence length at 128 tokens. These decisions have a significant impact on overall performance. Although the reported results in related work are SOTA or close to SOTA, these studies are representative of a common problem in the field, namely the fact that the important evaluation details are not always communicated properly (micro vs.~macro F1 score, token- vs.~span-based evaluation, and evaluating the model prediction on the first token of tokenized words from the input sequence vs.~on all tokens). \citet{scaria2023instructabsa} achieve SOTA F1 score of $0.928$ on Rest14 dataset and ATE task with IT. \citet{yang2021improving} report SOTA macro F1 of $0.863$ on ATP task with DeBERTa model \cite{he2020deberta}. 
The SOTA F1 score of $69.8$ on ACE05 for TC is achieved by \cite{wang-etal-2022-deepstruct}.

\subsection{Investigating the Effect of Scale}


We observed gains upon CM removal on a 7B parameters scale. Our experiments prompt the question of whether similar findings would apply to models of different parameter scales. More specifically, whether CM removal from a small CLM-based decoder would exhibit the same trend, and also whether MLM-based pre-training is more beneficial for success on SL tasks when the number of parameters, pre-training data and steps, and all other hyperparameters between decoders and encoders are equal. To investigate this, we consider two models of comparable size -- a small CLM-based decoder and an MLM-based encoder. We randomly initialize small LMs with four decoder or encoder blocks following RoBERTa-base architecture with a language modeling head on top and a newly initialized embedding matrix with the size of RoBERTa-base's vocabulary. We inherit all other RoBERTa-base hyperparameters and produce a small LM with 68M parameters. We use the RoBERTa-base tokenizer to pre-process the BookCorpus dataset \citep{zhu2015aligning} consisting of 74M sentences and use this dataset to pre-train small LMs. We train the small decoder and encoder with CLM and MLM, respectively, with AdamW \citep{loshchilov2017decoupled} optimizer, a cosine annealing learning rate scheduler, eight gradient accumulation steps, and bfloat16 precision with a batch size of 64 and a learning rate of $2\mathrm{e}{-4}$. We save the model weights immediately after random initialization and then at five equally spaced intervals throughout each epoch (51 checkpoints). The training continues for a fixed number of 10 epochs (roughly 200K steps). After pre-training, we load the weights of each checkpoint, replace the LM head with the SL head for the appropriate task, and fine-tune all parameters for five epochs on a task-specific training set with a batch size of 16. We fine-tune three variants: a pre-trained encoder, a pre-trained decoder with CM during fine-tuning, and a pre-trained decoder without CM. We report averages over five fine-tuning runs, with the last model from each run evaluated on the validation set. More details are provided in Appendix~\ref{sec:scale_exp_details}. 

\begin{figure}
\begin{center}
\includegraphics[width=0.5\columnwidth]{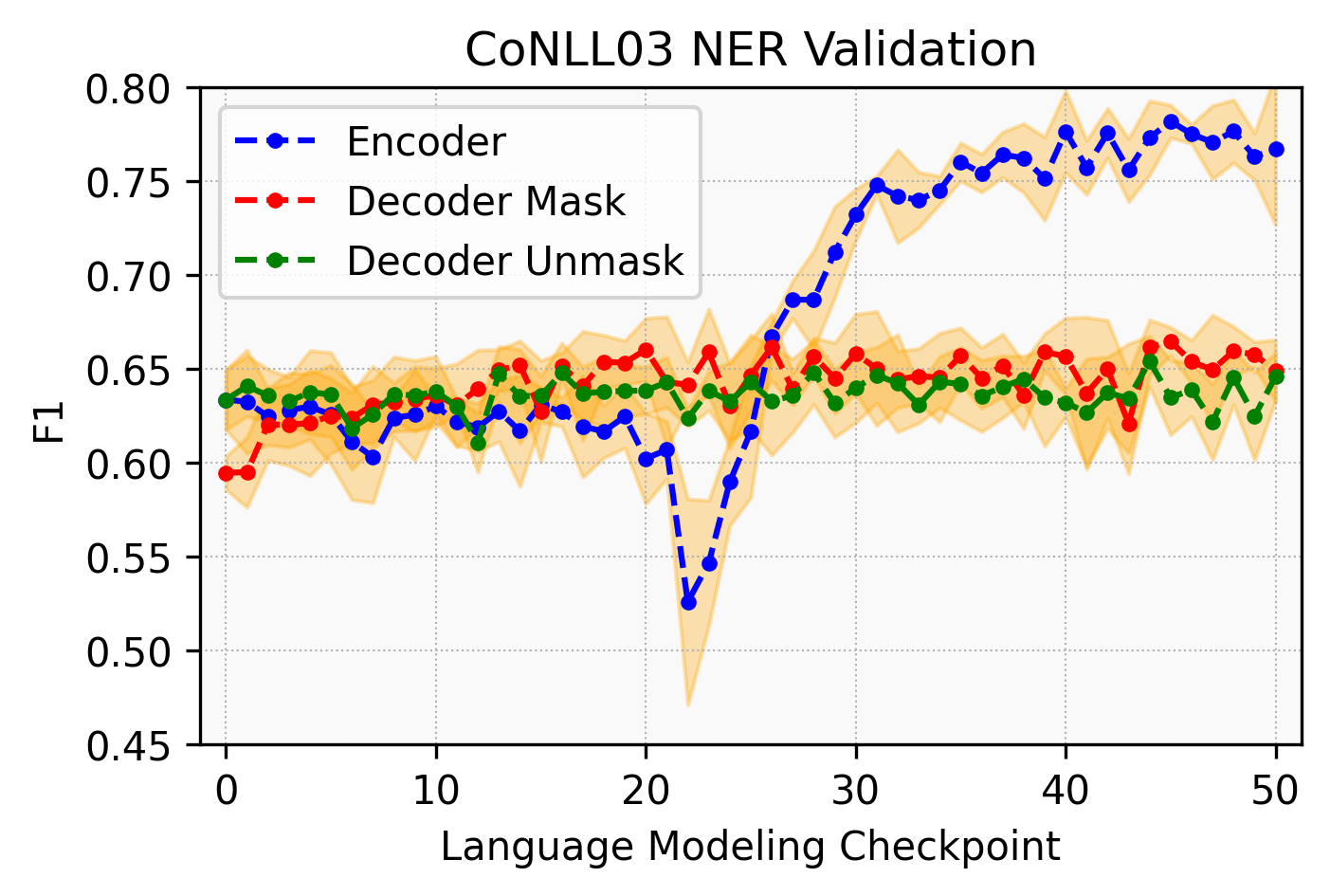}%
\includegraphics[width=0.5\columnwidth]{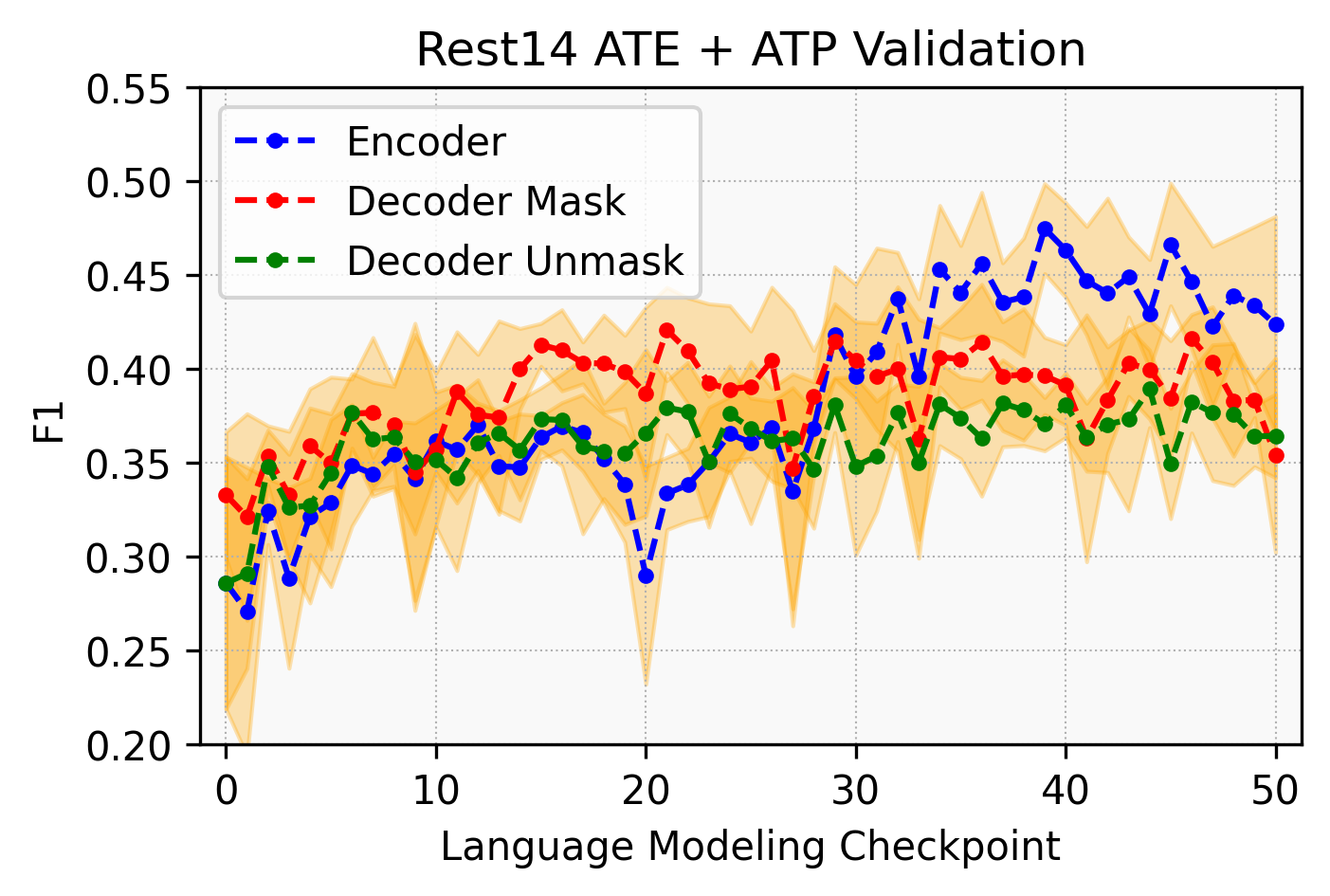}\\
\includegraphics[width=0.5\columnwidth]{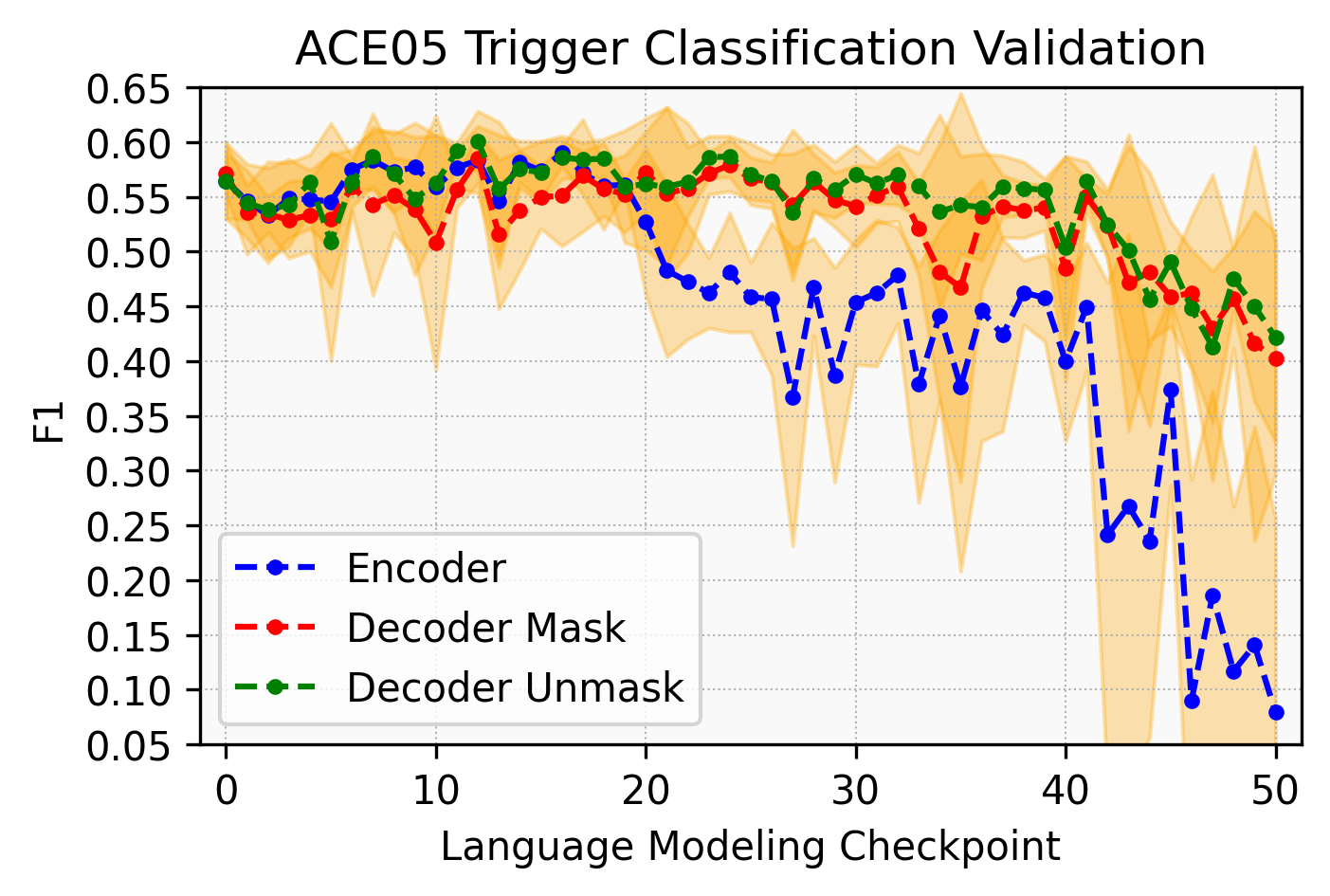}%
\includegraphics[width=0.5\columnwidth]{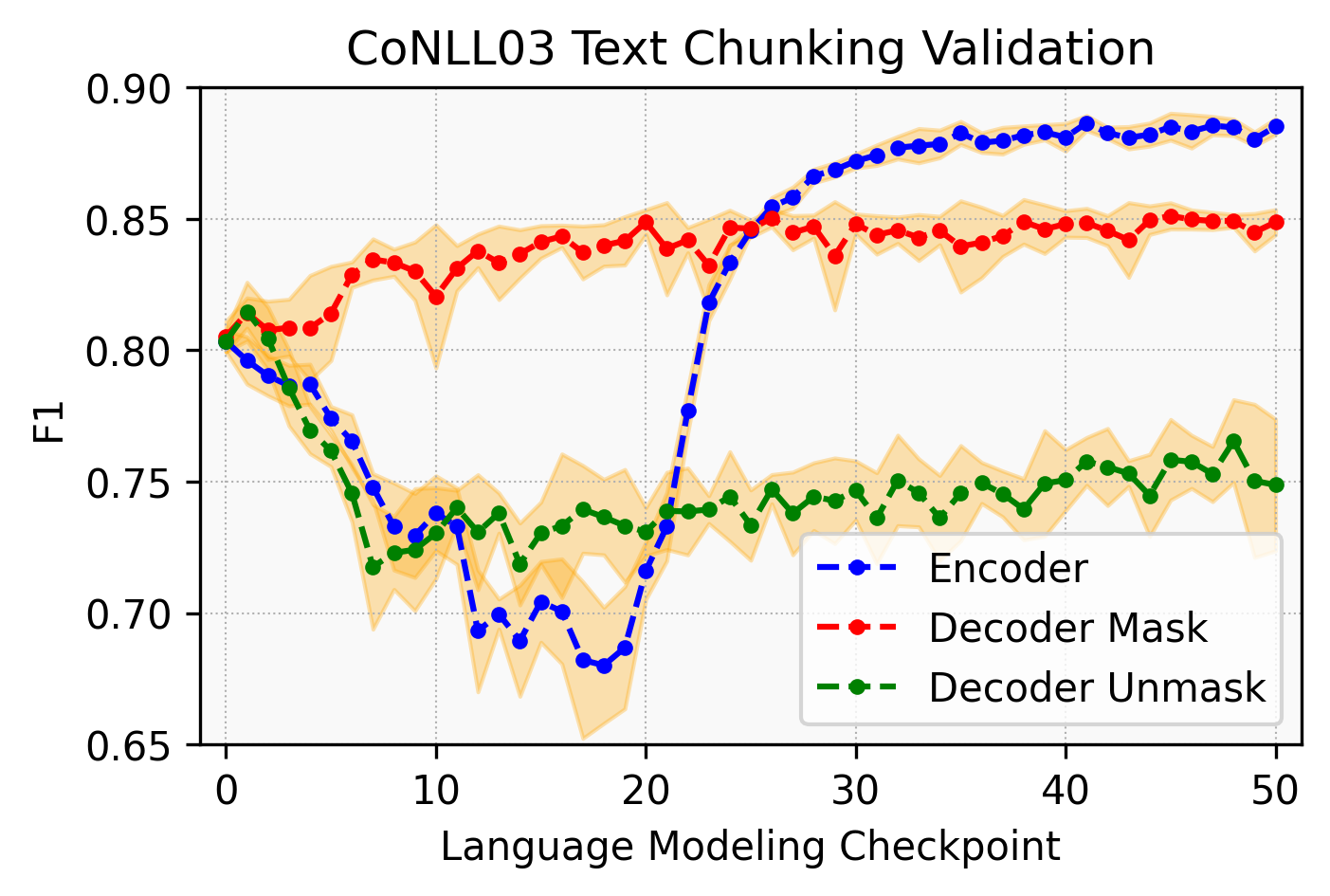}
\caption{Validation set micro F1 SL scores of small pre-trained MLM-based encoder and CLM-based decoder models after fine-tuning on SL tasks training data starting from particular LM checkpoint. Decoder Unmask: model pre-trained with CLM and a CM but fine-tuned without the CM. All results are averages over five runs. The shaded area represents the standard deviation.}
\label{fig:pretraining_roberta}
\end{center}
\end{figure}

The results in Figure~\ref{fig:pretraining_roberta} reveal that removing the CM on the 68M parameters scale produces no gains. On average, Decoder Unmask performs worse than Decoder Mask. Moreover, encoders struggle to keep up with decoders until around the 20th checkpoint (fourth pre-training epoch), when they start prevailing on all SL tasks except TC. MLM training for over four epochs drastically hurts performance on ACE05. The score RoBERTa-base achieves on ACE05 (cf.~Table~\ref{tab:comparison}) is close to the best small encoder. 
The gains from continued MLM training on ACE05 become less important than the overall number of model parameters. This can be explained by the fact that RoBERTa-base and RoBERTa-large were trained for 500k steps \citep{liu2019roberta}, and RoBERTa-large is drastically better on ACE05.


\section{Conclusion}

The landscape of PLMs shifted from encoder to decoder dominance in various NLU tasks. 
While large decoder models show remarkable performance even without fine-tuning, challenges persist in IE tasks, typically formulated as sequence labeling. 
In a series of experiments, we showed that LLMs can yield performance competitive with SOTA on sequence labeling IE tasks when the causal mask is removed from specific decoder blocks.
Future work should investigate CM removal for individual layers, analyze task-specific differences in CM removal, and investigate the options for predicting the impact of CM removal without requiring extensive fine-tuning. Promising alternatives to trying out all the unmasking configurations could involve leveraging dynamic programming or estimating a network's final performance without retraining, for example, using methods that rely on the Fisher information, similar to the FIT method by \citet{zandonati2022fit}. Additionally, performance estimation for CM removal from a layer could be improved using representation similarity analysis methods such as centered kernel alignment \cite{kornblith2019similarity}. These approaches might reduce the number of unmasking configurations that need to be evaluated to find the best one. Comparisons with encoders stress the significance of architecture and scale in SL tasks. A noticeable gap exists between small-scale encoder and decoder models, where decoder models do not benefit from CM removal at a small scale. 

\section{Limitations}

In our experiments, we ensure reliability by averaging performance scores over five runs with different random seeds. Increasing the sample size for averaging would enhance reliability further. However, maintaining fixed learning rates and other hyperparameters across experiments might have led to suboptimal adaptation for sequence labeling tasks. Exploring additional unmasking configurations could provide valuable insights and potential improvements. While numerous open LLMs are available, we focus solely on Llama2 and Mistral, both with 7B parameters. Enhancing prompt templates used for instruction tuning could boost overall instruction tuning performance. Given more computing resources, experimenting with larger models would be feasible. Additionally, our experiments were limited to English-language datasets. Extending the analysis to sequence labeling tasks in other languages and incorporating more diverse datasets could yield further insights. Finally, the main limitation of our work is the fact that one needs to try out all the possible unmasking configurations to find the best one on the validation set. Trying out all the possible configurations requires extensive fine-tuning.

\bibliography{anthology,custom}

\appendix

\section{Appendix} \label{sec:appendix}

\subsection{Experimental Setup Details} \label{sec:exp_setup_details}

Since RoBERTa requires tokenization with added prefix space, we enforce the same for open LLMs' tokenizers. We use the end-of-sequence token for the models with no pre-set padding token. The cross-entropy loss is adjusted to consider only the first token of each tokenized word from the input sequence. We report all results as averages of five runs and use $\mathit{seeds}=\{120,121,122,123,124\}$.

For IT experiments, we generate the outputs for evaluation with default generation settings for the Llama2-7B model. To speed up the generation, we decrease the total maximum length of the input instruction prompt combined with newly generated tokens to 1024. The same generation config is used for Mistral-7B.

\subsection{Training Small Language Models Details} \label{sec:scale_exp_details}

We use a RoBERTa-base tokenizer with added prefix space. Tokenized BookCorpus sentences are grouped to form chunks of size 512 for either MLM-based pre-training of small encoder or CLM-based pre-training of small decoder LM. MLM probability is set at $0.15$. Models are trained with AdamW. Its parameters are fixed to $\beta_1=0.9, \beta_2=0.95, \epsilon=1\mathrm{e}{-5}, \lambda=0.1$ and we apply gradient clipping to $1.0$. We pre-train both models with the same seed. We fine-tune for SL tasks five times and use $\mathit{seeds}=\{120,121,122,123,124\}$.

\subsection{Instruction Tuning Details} \label{sec:appendix_it}

\begin{table*}
    \centering
    \begin{adjustbox}{width=0.9\textwidth}
    \begin{tabular}{p{2.5cm}  p{4.9cm}  p{4.9cm}}
    \toprule
         \multicolumn{1}{c}{Task (dataset)} & \multicolumn{1}{c}{Instruction example for training} & \multicolumn{1}{c}{Instruction example for evaluation} \\
         \midrule
         \multicolumn{1}{c}{\multirow{3}{*}{NER (CoNLL03)}} & \scriptsize
         \#\#\# Instruction: 

please extract named entities and their type from the input sentence, all entity types are in options 

\#\#\# Options:

person, location, organization, miscellaneous

\#\#\# Sentence:

" What we have to be extremely careful of is how other countries are going to take Germany 's lead , " Welsh National Farmers ' Union ( NFU ) chairman John Lloyd Jones said on BBC radio .

\#\#\# Response:

Germany:location;Welsh National Farmers ' Union:organization;NFU:organization;John Lloyd Jones:person;BBC radio:organization & \scriptsize
         \#\#\# Instruction: 

please extract named entities and their type from the input sentence, all entity types are in options 

\#\#\# Options:

person, location, organization, miscellaneous

\#\#\# Sentence:

" What we have to be extremely careful of is how other countries are going to take Germany 's lead , " Welsh National Farmers ' Union ( NFU ) chairman John Lloyd Jones said on BBC radio .

\#\#\# Response:  \\\midrule

\multicolumn{1}{c}{\multirow{3}{*}{\shortstack{ATE+ATP \\ (Rest14)}}} & \scriptsize \#\#\# Instruction:

please extract aspect terms and their polarity from the input sentence, all polarity types are in options 

\#\#\# Options:

positive, negative, neutral, conflict

\#\#\# Sentence:

The lobster sandwich is \$ 24 and although it was good it was not nearly enough to warrant that price .

\#\#\# Response:

lobster sandwich:conflict;price:negative & \scriptsize \#\#\# Instruction:

please extract aspect terms and their polarity from the input sentence, all polarity types are in options 

\#\#\# Options:

positive, negative, neutral, conflict

\#\#\# Sentence:

The lobster sandwich is \$ 24 and although it was good it was not nearly enough to warrant that price .

\#\#\# Response: \\\midrule
\multicolumn{1}{c}{\multirow{3}{*}{\shortstack{Trigger Classification \\ (ACE05)}}} & \scriptsize \#\#\# Instruction: 

please extract events and their types from the input sentence, all event types are in options 

\#\#\# Options:

merge organization, start organization, declare bankruptcy, end organization, grant pardon, extradite, execute, impose fine, conduct trial hearing, issue sentence, file appeal, convict, file lawsuit, release on parole, arrest and send to jail, charge and indict, acquit, participate in protest or demonstration, attack, contact via written or telephone communication, meet, start position, elect, end position, nominate, transfer ownership, transfer money, marry, divorce, be born, die, sustain injury, transport

\#\#\# Sentence:

In his previous letter home , Apache pilot Joe Bruhl did n't tell his family the full details about his first combat mission .

\#\#\# Response:

tell:contact via written or telephone communication;combat:attack & \scriptsize \#\#\# Instruction: 

please extract events and their types from the input sentence, all event types are in options 

\#\#\# Options:

merge organization, start organization, declare bankruptcy, end organization, grant pardon, extradite, execute, impose fine, conduct trial hearing, issue sentence, file appeal, convict, file lawsuit, release on parole, arrest and send to jail, charge and indict, acquit, participate in protest or demonstration, attack, contact via written or telephone communication, meet, start position, elect, end position, nominate, transfer ownership, transfer money, marry, divorce, be born, die, sustain injury, transport

\#\#\# Sentence:

In his previous letter home , Apache pilot Joe Bruhl did n't tell his family the full details about his first combat mission .

\#\#\# Response: \\\midrule
\multicolumn{1}{c}{\multirow{3}{*}{\shortstack{Text Chunking \\ (CoNLL03)}}} & \scriptsize
\#\#\# Instruction: 

please extract chunks and their type from the input sentence, all chunk types are in options 

\#\#\# Options:

noun phrase, verb phrase, prepositional phrase, adverb phrase, subordinated clause, adjective phrase, particles, conjunction phrase, interjection, list marker, unlike coordinated phrase

\#\#\# Sentence:

Rare Hendrix song draft sells for almost \$ 17,000 .

\#\#\# Response:

Rare Hendrix song draft:noun phrase;sells:verb phrase;for:prepositional phrase;almost \$ 17,000:noun phrase & \scriptsize \#\#\# Instruction: 

please extract chunks and their type from the input sentence, all chunk types are in options 

\#\#\# Options:

noun phrase, verb phrase, prepositional phrase, adverb phrase, subordinated clause, adjective phrase, particles, conjunction phrase, interjection, list marker, unlike coordinated phrase

\#\#\# Sentence:

Rare Hendrix song draft sells for almost \$ 17,000 .

\#\#\# Response: \\
    \bottomrule
    \end{tabular}
    \end{adjustbox}
    \caption{Instruction tuning examples from the training sets of the four SL datasets.}
\end{table*}

\end{document}